\documentclass[conference,letterpaper]{IEEEtran}
\IEEEoverridecommandlockouts

\usepackage{cite}
\usepackage{amsmath,amssymb,amsfonts}
\usepackage{algorithmic}
\usepackage{graphicx}
\usepackage{textcomp}
\usepackage[table]{xcolor}
\usepackage{booktabs}
\usepackage{multirow}
\usepackage{url}
\usepackage[hidelinks]{hyperref}
\usepackage{tikz}
\usepackage{geometry}
\usepackage{stfloats}
\usepackage{microtype}

\definecolor{methodgray}{RGB}{245,245,245}   
\definecolor{fitblue}{RGB}{233,242,255}      
\usepackage{caption}
\usepackage{subcaption}

\graphicspath{{figs/}}

\def\BibTeX{{\rm B\kern-.05em{\sc i\kern-.025em b}\kern-.08em
    T\kern-.1667em\lower.7ex\hbox{E}\kern-.125emX}}

\newcommand{\safeincludegraphics}[2][]{%
  \IfFileExists{#2}{\includegraphics[#1]{#2}}{%
    \fbox{\parbox[c][0.22\textheight][c]{0.92\linewidth}{\centering Missing figure: \texttt{#2}}}%
  }%
}

\begin{document}

\title{Language as Prior, Vision as Calibration: Metric Scale Recovery for Monocular Depth Estimation}
\author{
\IEEEauthorblockN{
Mingxia Zhan\IEEEauthorrefmark{1},
Li Zhang\IEEEauthorrefmark{1},
Beibei Wang\IEEEauthorrefmark{2},
Yingjie Wang\IEEEauthorrefmark{3},
Zenglin Shi\IEEEauthorrefmark{1}
}
\IEEEauthorblockA{\IEEEauthorrefmark{1}Hefei University of Technology, Hefei, Anhui, China}
\IEEEauthorblockA{\IEEEauthorrefmark{2}Institute of Artificial Intelligence, Hefei Comprehensive National Science Center, Hefei, Anhui, China}
\IEEEauthorblockA{\IEEEauthorrefmark{3}University of Science and Technology of China, Hefei, Anhui, China}

\thanks{Mingxia Zhan is with the Hefei University of Technology (HFUT), Hefei, Anhui, China (email: mxzhan@mail.hfut.edu.cn).}%
\thanks{Li Zhang is with the Hefei University of Technology (HFUT), Hefei, Anhui, China (email: lizhang@hfut.edu.cn).}%
\thanks{Beibei Wang is with the Institute of Artificial Intelligence, Hefei Comprehensive National Science Center, Hefei, Anhui, China (email: wbb@iai.ustc.edu.cn).}%
\thanks{Yingjie Wang is with the University of Science and Technology of China (USTC), Hefei, Anhui, China (email: yingjiewang@mail.ustc.edu.cn).}%
\thanks{Zenglin Shi is with the Hefei University of Technology (HFUT), Hefei, Anhui, China (email: zenglin.shi@hfut.edu.cn).}%
}

\maketitle

\begin{abstract}
Relative-depth foundation models transfer well, yet monocular metric depth remains ill-posed due to unidentifiable global scale and heightened domain-shift sensitivity. Under a frozen-backbone calibration setting, we recover metric depth via an image-specific affine transform in inverse depth and train only lightweight calibration heads while keeping the relative-depth backbone and the CLIP text encoder fixed. Since captions provide coarse but noisy scale cues that vary with phrasing and missing objects, we use language to predict an uncertainty-aware envelope that bounds feasible calibration parameters in an unconstrained space, rather than committing to a text-only point estimate. We then use pooled multi-scale frozen visual features to select an image-specific calibration within this envelope. During training, a closed-form least-squares oracle in inverse depth provides per-image supervision for learning the envelope and the selected calibration. Experiments on NYUv2 and KITTI improve in-domain accuracy, while zero-shot transfer to SUN-RGBD and DDAD demonstrates improved robustness over strong language-only baselines.
\end{abstract}


\section{Introduction}
Monocular depth estimation, which infers scene depth from a single RGB image, is a key component in 3D perception pipelines~\cite{thorough_review,Li2023}. However, metric scale is inherently unobservable from a single view, making metric depth estimation an ill-posed problem. Although modern foundation models yield robust relative depth estimates that transfer effectively across datasets~\cite{yang2024depthanythingv2}, the conversion to absolute metric depth remains contingent upon resolving a global scale ambiguity, which can be accurately modeled as an affine transformation in inverse depth space~\cite{eigen2014depth}. This ambiguity becomes especially problematic under domain shift, where even minor calibration errors can lead to significant scale drift and degrade downstream task performance.

Existing solutions for metric depth estimation typically fall into two categories. The first paradigm leverages auxiliary sensing or geometric cues (e.g., sparse range measurements, multi-view constraints, or acquisition-time metadata) to anchor absolute scale at capture or inference time~\cite{lin2025prompting}, but these signals are often unavailable for unconstrained in-the-wild imagery. The second paradigm learns metric scale through supervised training, typically by training on datasets with ground-truth depth conditioned on known camera intrinsics~\cite{bhat2023zoedepth}. However, the learned scale estimates often exhibit strong dependence on the training domain and sensor configuration, leading to poor generalization under shifts in depth range, camera parameters, or scene statistics. As modern vision systems increasingly operate on diverse and heterogeneous RGB collections without consistent sensing conditions, the disconnect between widely available images and reliable metric grounding remains a critical bottleneck~\cite{zeng2024rsa}. 

To bridge this gap without additional sensors, recent work has explored vision–language pretraining as a source of scene-level priors for metric scale. Intuitively, language can convey coarse but informative world-scale hints: reference to object categories (e.g., \textit{car} vs \textit{toy car}) and scene types (e.g., \textit{indoor rooms} vs. \textit{urban streets}) correlates with typical object sizes and camera-to-scene distances, which constrains the plausible order of magnitude of metric depth. Large-scale vision--language models trained on image-text pairs, such as CLIP~\cite{radford2021learning}, implicitly encode knowledge about object semantics with their typical sizes, and contextual scene structures that inform coarse metric scale. Prior studies leverage CLIP-style representations for depth estimation, including zero-shot depth via CLIP responses~\cite{zhang2022can} and prompting or task conditioning for improved transfer~\cite{auty2023learning}. In parallel, WorDepth~\cite{zeng2024wordepth} leverages captions as a probabilistic prior to constrain plausible 3D scene structure, while RSA~\cite{zeng2024rsa} uses language to regress global scale and shift for calibrating a frozen relative-depth backbone to metric depth under an affine-in-inverse-depth model. 

Despite these advances, textual cues remain inherently coarse and often incomplete as a source of metric information. First, automatically generated captions may vary across paraphrases or omit objects critical for scale inference. Second, semantically similar descriptions can correspond to scenes with markedly different layouts, object proportions, and depth ranges. These ambiguities make language-only calibration brittle under domain shift.

This motivates a role-aware design. Vision provides geometric structure, while language supplies a coarse, uncertainty-aware prior over global metric calibration. We study metric grounding under a frozen-backbone protocol: both the relative-depth backbone and the CLIP text encoder are kept fixed, and only lightweight calibration heads are trained. This setup preserves the backbone's transferable relative geometry and isolates the generalization of global calibration under domain shift. Building on this protocol, we propose a language-guided depth calibration framework that operates atop a frozen relative-depth backbone. Specifically, we introduce two lightweight and complementary heads. A caption-conditioned head predicts an uncertainty-aware envelope that bounds feasible scale-and-shift parameters in an unconstrained calibration space. A vision-conditioned selector aggregates frozen multi-scale visual features to choose an image-specific calibration within this envelope. During training, we compute a closed-form per-image least-squares calibration in inverse depth as an oracle signal to supervise the calibration heads. We train the envelope predictor and selector jointly with metric depth supervision. Trained on NYUv2 and KITTI, our calibration improves in-domain performance and shows stronger zero-shot robustness on SUN-RGBD and DDAD than language-guided baselines~\cite{silberman2012indoor,geiger2013vision,song2015sun}.

In summary, this work makes the following contributions:

\noindent$\bullet$ We propose an envelope-and-selection framework that treats language as a bounded prior and vision as the per-image calibrator on a frozen backbone.

\noindent$\bullet$ We design a lightweight calibration layer that enables robust, per-image estimation of scale/shift parameters without fine-tuning the backbone.

\noindent$\bullet$ We validate the approach on NYUv2 and KITTI and evaluate zero-shot transfer to SUN-RGBD and DDAD, and show improved robustness to domain shift over language-guided baselines.

\begin{figure*}[t] \centering \safeincludegraphics[width=0.95\textwidth,height=0.45\textheight,keepaspectratio]{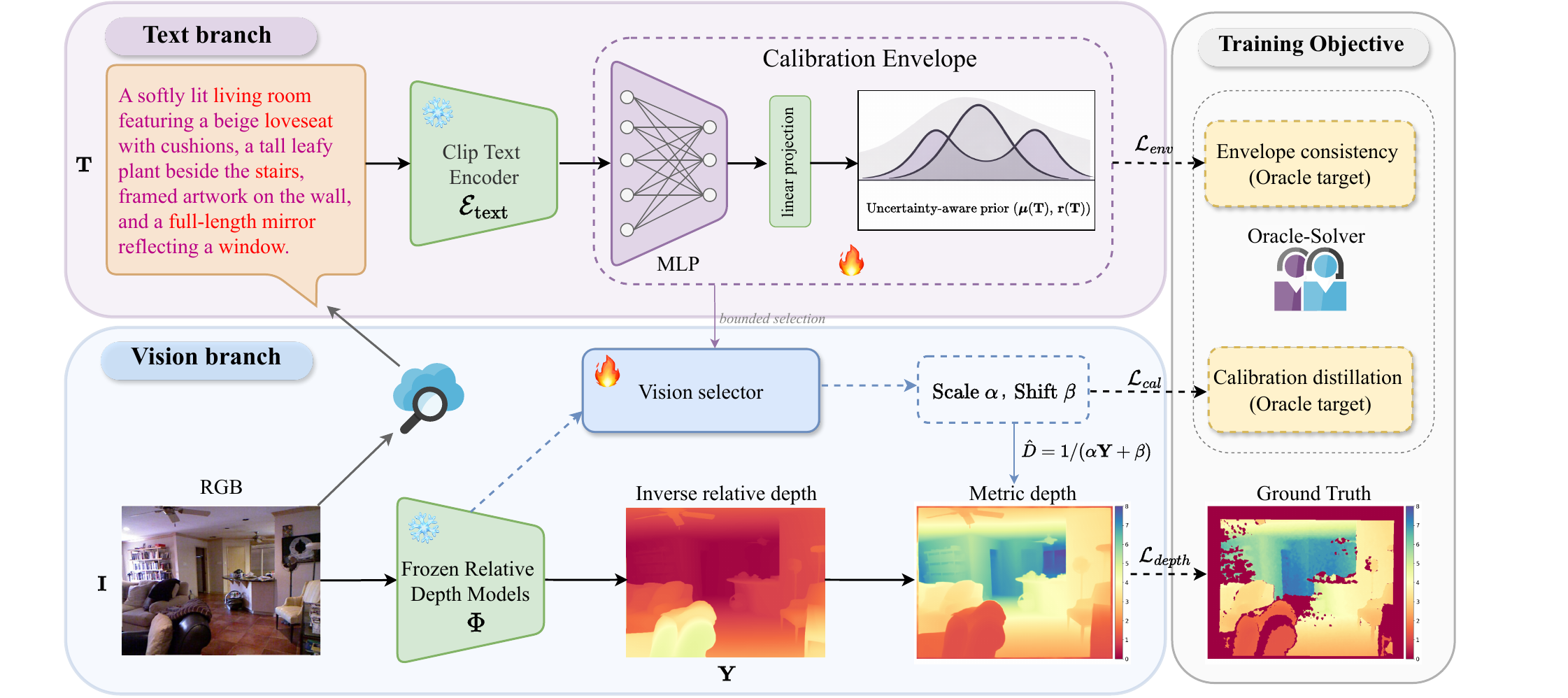} \caption{Framework overview. Given an image $\mathbf{I}$, a frozen backbone $\Phi$ outputs inverse relative depth $\mathbf{Y}$ and multi-scale features. The caption $\mathbf{T}$ is auto-generated by LLaVA v1.6. A frozen CLIP text encoder maps $\mathbf{T}$ to an uncertainty-aware envelope $(\boldsymbol{\mu}(\mathbf{T}),\mathbf{r}(\mathbf{T}))$, and a vision-conditioned selector predicts an image-specific calibration within this envelope. Training uses an online inverse-depth least-squares Oracle-Solver to provide per-image targets for envelope consistency, calibration distillation, and depth reconstruction losses.}
 \label{fig:framework} 
\end{figure*}

\section{Method}

\subsection{Preliminary}
We assume a frozen relative-depth backbone and recover metric depth via a global affine calibration in inverse depth. We train on triplets $(\mathbf{I},\mathbf{T},\mathbf{D}_{\mathrm{gt}})$, where $\mathbf{T}$ is generated automatically at test time. Given $\mathbf{I}$, a frozen relative-depth backbone $\Phi$ predicts inverse relative depth $\mathbf{Y}=\Phi(\mathbf{I})$.
We predict an unconstrained 2D calibration vector $\tilde{\boldsymbol{\theta}}=(\tilde{\alpha},\tilde{\beta})$ and map it to
\begin{equation}
\alpha=\mathrm{softplus}(\tilde{\alpha}),\qquad
\beta=\beta_{\min}+(\beta_{\max}-\beta_{\min})\cdot\sigma(\tilde{\beta}),
\label{eq:theta_map}
\end{equation}
where $\sigma(\cdot)$ is the sigmoid function and $\beta_{\min}<\beta_{\max}$ are fixed constants.
We then recover metric depth by an affine calibration in inverse depth:
\begin{equation}
\widehat{\mathbf{D}}(x)=\frac{1}{\max\!\left(\alpha\,\mathbf{Y}(x)+\beta,\ \varepsilon\right)}.
\label{eq:metric}
\end{equation}

\subsection{Overview}
Our goal is to predict the global calibration from complementary cues of language and vision.
We factorize the unconstrained calibration vector into a caption-conditioned envelope and an image-conditioned offset:
\begin{equation}
\tilde{\boldsymbol{\theta}}
=
\boldsymbol{\mu}(\mathbf{T})+\mathbf{r}(\mathbf{T})\odot\boldsymbol{\delta}(\mathbf{I}),
\label{eq:compose}
\end{equation}
where $\boldsymbol{\mu}(\mathbf{T})\in\mathbb{R}^{2}$ is the envelope center,
$\mathbf{r}(\mathbf{T})\in\mathbb{R}_{+}^{2}$ is the envelope radius,
and $\boldsymbol{\delta}(\mathbf{I})\in[-1,1]^2$ is an instance-specific offset predicted from frozen visual features.
Here $\odot$ denotes element-wise multiplication: language predicts $(\boldsymbol{\mu}(\mathbf{T}),\mathbf{r}(\mathbf{T}))$ as a feasible set in the unconstrained calibration space, while vision predicts $\boldsymbol{\delta}(\mathbf{I})$ to select an image-specific calibration within this set.
We obtain $(\alpha,\beta)$ via Eq.~(\ref{eq:theta_map}) and compute $\widehat{\mathbf{D}}$ by Eq.~(\ref{eq:metric}).

\subsection{Caption-conditioned Calibration Envelope}
Captions provide coarse scene-level cues for metric scale but are noisy.
We model language as an uncertainty-aware prior by predicting an envelope $(\boldsymbol{\mu}(\mathbf{T}),\mathbf{r}(\mathbf{T}))$.

We encode the caption with a frozen CLIP text encoder $\mathbf{z}=\mathcal{E}_{\mathrm{text}}(\mathbf{T})$,
and map it to $\mathbf{h}=\mathrm{MLP}(\mathbf{z})$ using a lightweight 3-layer MLP trunk (hidden width 256; ReLU).
Two linear heads output
\begin{equation}
\boldsymbol{\mu}(\mathbf{T})=\mathbf{W}_{\mu}\mathbf{h}+\mathbf{b}_{\mu},\qquad
\mathbf{r}(\mathbf{T})=\mathrm{softplus}(\mathbf{W}_{r}\mathbf{h}+\mathbf{b}_{r}),
\end{equation}
where $\mathbf{r}(\mathbf{T})$ is bounded element-wise by a fixed upper limit $\mathbf{r}_{\max}$ (implemented by clamping).
We train the envelope with the following penalty (target defined in Sec.~\ref{sec:train}):
\begin{equation}
\mathcal{L}_{\mathrm{env}}
=
\sum_{k=1}^{2}
\mathrm{softplus}\!\left(\left|\tilde{\theta}_k^{\star}-\mu_k(\mathbf{T})\right|-r_k(\mathbf{T})\right).
\label{eq:envloss}
\end{equation}

\subsection{Vision-conditioned Calibration Selection}
Language alone cannot determine image-specific calibration. We therefore predict a bounded offset
$\boldsymbol{\delta}(\mathbf{I})$ from frozen backbone features.
We extract a 4-scale feature pyramid from $\Phi$ at effective strides $\{4,8,16,32\}$ (i.e., resolutions $\{H/4,H/8,H/16,H/32\}$ w.r.t.\ the input),
denoted as $\{\mathbf{F}_{\ell}\}_{\ell=1}^{4}$. We apply global average pooling and concatenate:
\begin{equation}
\mathbf{s}_{\ell}=\mathrm{GAP}(\mathbf{F}_{\ell}),\qquad
\mathbf{s}=[\mathbf{s}_{1};\ldots;\mathbf{s}_{4}].
\end{equation}
A lightweight 3-layer MLP $g_{\mathrm{sel}}(\cdot)$ (hidden width 256; ReLU) regresses the offset:
\begin{equation}
\boldsymbol{\delta}(\mathbf{I})=\tanh\!\left(g_{\mathrm{sel}}(\mathbf{s})\right).
\end{equation}

\subsection{Training Strategy}
\label{sec:train}
We optimize only $(\boldsymbol{\mu}(\cdot),\mathbf{r}(\cdot))$ and $g_{\mathrm{sel}}(\cdot)$, with $\Phi$ and $\mathcal{E}_{\mathrm{text}}$ fixed. Let $\Omega=\{x:\mathbf{D}_{\mathrm{gt}}(x)>0\}$ be the valid pixel set and define
\begin{equation}
\mathbf{Y}_{\mathrm{gt}}(x)=\frac{1}{\max\!\left(\mathbf{D}_{\mathrm{gt}}(x),\ \varepsilon\right)}.
\end{equation}

\textbf{Online closed-form oracle.}
For each training image, we compute an oracle calibration $(\alpha_{\mathrm{ls}},\beta_{\mathrm{ls}})$ by least-squares fitting in inverse depth:
\begin{equation}
(\alpha_{\mathrm{ls}},\beta_{\mathrm{ls}})
=
\arg\min_{\alpha,\beta}
\frac{1}{|\Omega|}
\sum_{x\in\Omega}
\left(\alpha\,\mathbf{Y}(x)+\beta-\mathbf{Y}_{\mathrm{gt}}(x)\right)^2.
\label{eq:teacher_ls}
\end{equation}
This regression has a closed-form solution. Let $y=\mathbf{Y}(x)$ and $g=\mathbf{Y}_{\mathrm{gt}}(x)$ for $x\in\Omega$.
The solution is $\alpha_{\mathrm{ls}}=\mathrm{Cov}(y,g)/\max\!\left(\mathrm{Var}(y),\varepsilon\right)$ and
$\beta_{\mathrm{ls}}=\mathbb{E}[g]-\alpha_{\mathrm{ls}}\mathbb{E}[y]$.
For stability, we clamp $\alpha_{\mathrm{ls}}\leftarrow\mathrm{clip}(\alpha_{\mathrm{ls}},\varepsilon,\alpha_{\max})$ and
$\beta_{\mathrm{ls}}\leftarrow\min\!\big(\max(\beta_{\mathrm{ls}},\beta_{\min}),\beta_{\max}\big)$, where $\alpha_{\max}$ is a fixed large constant (we use the same $\alpha_{\max}$ for all datasets).
We fit the oracle in inverse depth, which yields a stable linear calibration.

We form the stop-gradient target $\boldsymbol{\theta}^{\star}=(\alpha_{\mathrm{ls}},\beta_{\mathrm{ls}})$.
For the envelope penalty in Eq.~(\ref{eq:envloss}), we define the corresponding unconstrained target
$\tilde{\boldsymbol{\theta}}^{\star}=(\tilde{\alpha}^{\star},\tilde{\beta}^{\star})$ with
$\tilde{\alpha}^{\star}=\mathrm{softplus}^{-1}(\alpha_{\mathrm{ls}})$ and
$\tilde{\beta}^{\star}=\mathrm{logit}(p)$, where $p=(\beta_{\mathrm{ls}}-\beta_{\min})/(\beta_{\max}-\beta_{\min})$.
In implementation, we use numerically stable inverse mappings and clamp $p$ to $[\varepsilon,1-\varepsilon]$ to avoid overflow.

\textbf{Unified end-to-end objective.}
We compute $\widehat{\mathbf{D}}$ from Eq.~(\ref{eq:compose})--(\ref{eq:metric}) and optimize:
\begin{equation}
\mathcal{L}
=
\mathcal{L}_{\mathrm{depth}}
+
\lambda_{\mathrm{env}}\mathcal{L}_{\mathrm{env}}
+
\lambda_{r}\|\mathbf{r}(\mathbf{T})\|_{1}
+
\lambda_{\mathrm{cal}}\|(\alpha,\beta)-\mathrm{sg}(\boldsymbol{\theta}^{\star})\|_{1},
\label{eq:unified_loss}
\end{equation}
where $\mathrm{sg}(\cdot)$ denotes stop-gradient and
\begin{equation}
\mathcal{L}_{\mathrm{depth}}
=
\frac{1}{|\Omega|}\sum_{x\in\Omega}\left|\widehat{\mathbf{D}}(x)-\mathbf{D}_{\mathrm{gt}}(x)\right|.
\end{equation}
Here $\mathcal{L}_{\mathrm{cal}}$ distills the per-image oracle calibration parameters, while $\mathcal{L}_{\mathrm{depth}}$ directly aligns the final metric depth with the evaluation target.

\section{Experiments}

\subsection{Experimental setup}

\textbf{Datasets.}
We use NYUv2~\cite{silberman2012indoor} (indoor), KITTI~\cite{eigen2014depth,geiger2013vision} (outdoor), and additionally evaluate on SUN-RGBD~\cite{song2015sun} and DDAD~\cite{guizilini20203d}. NYUv2 consists of $480\times640$ indoor images with metric depth, using the standard train/test split. KITTI consists of $352\times1216$ outdoor driving images; we adopt the Eigen split~\cite{eigen2014depth} and follow common practice to exclude test frames without valid LiDAR ground truth. SUN-RGBD and DDAD are used for cross-dataset evaluation and provide 5{,}050 test images and 3{,}950 validation images, respectively.

\textbf{Evaluation Metrics.}
We report standard metric depth metrics~\cite{eigen2014depth}:
Abs Rel, RMSE, $\mathrm{RMSE}_{\log}$, $\log_{10}$, and threshold accuracies $\delta < 1.25^k$.

\textbf{Depth Models.}
We evaluate two representative depth foundation models: DPT~\cite{ranftl2021vision} and DepthAnything~\cite{yang2024depthanythingv2}.
We use DPT-Hybrid fine-tuned on NYUv2 and KITTI, and DepthAnything-Small (24.8M parameters). Following our problem setting, we adopt a unified global scale-and-shift calibration to convert relative depth to metric depth, instead of using dataset-specific pixel-wise metric decoders (e.g., ZoeDepth-style~\cite{bhat2023zoedepth}). All baselines are re-implemented under this protocol, which reflects deployment scenarios with frozen backbones and limited metric supervision.

\begin{figure*}[t]
\centering
\safeincludegraphics[width=0.90\textwidth,height=0.30\textheight,keepaspectratio]{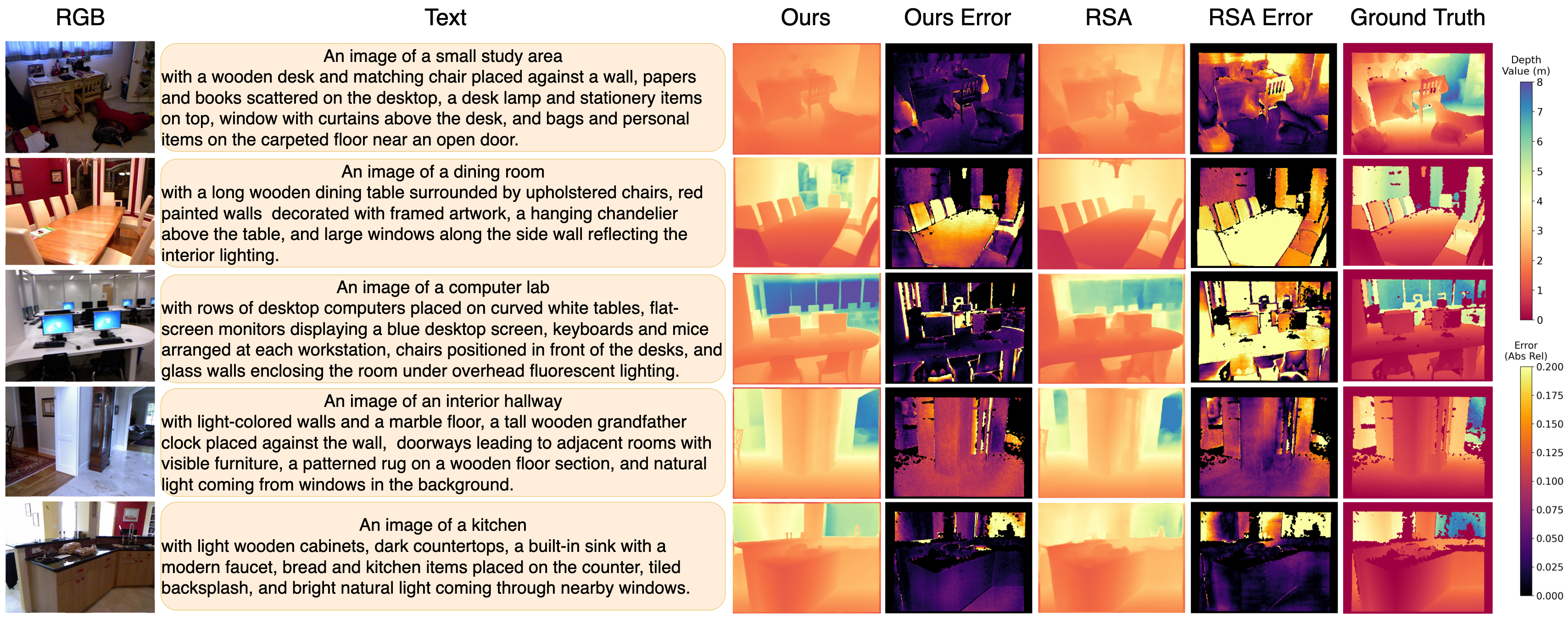}
\captionof{figure}{Qualitative results on NYUv2. Depth predictions and absolute error maps of our method and RSA are shown for comparison. Both approaches build upon the same DPT backbone and differ only in the calibration strategy. While the overall depth structures remain similar, our method consistently reduces estimation errors across indoor scenes, particularly in regions where scale mismatch is visually prominent.}
\label{fig:qual_nyu}

\vspace{2pt}

\safeincludegraphics[width=0.90\textwidth,height=0.50\textheight,keepaspectratio]{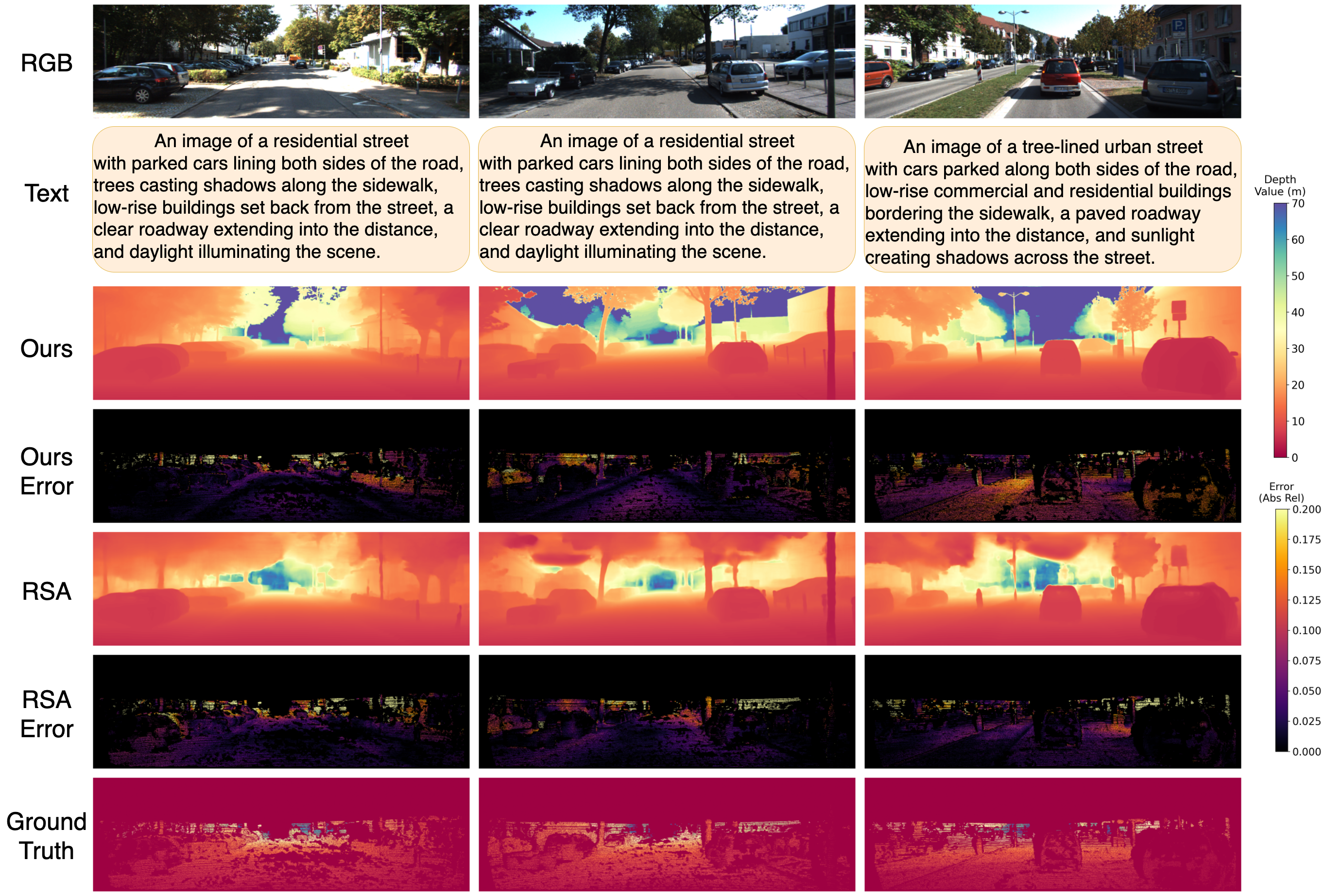}
\captionof{figure}{Qualitative results on KITTI under the same setting as Figure~\ref{fig:qual_nyu}. While preserving similar scene geometry, our method achieves improved metric scale consistency and lower depth errors, especially in long-range outdoor regions.}
\label{fig:qual_kitti}
\end{figure*}

\subsection{Implementation Details}
\textbf{Caption protocol.}
Since standard depth benchmarks do not provide paired text, we generate captions using LLaVA v1.6~\cite{liu2024improved}.
We pre-generate $K{=}15$ one-sentence captions per image using LLaVA v1.6 with Vicuna and Mistral checkpoints, with 5 prompt templates per model. During training, we randomly sample one caption per iteration. During testing, we generate one caption per image using a fixed prompt template and decoding configuration for reproducibility. To reduce sensitivity to any single caption generator, we train with multi-caption augmentation and explicitly evaluate caption-induced variance in Table~\ref{tab:text_sensitivity_params}.

\textbf{Hyperparameters.}
We use AdamW~\cite{loshchilov2019decoupled} and a cosine learning-rate schedule,
where the learning rate decays from $3\times10^{-5}$ to $1\times10^{-5}$ over 50 epochs. We use a batch size of 8 and set the envelope radius upper bound to $r_{\max}=3$ in all experiments. For numerical stability, we set $\varepsilon$ to $10^{-6}$. The regularization weight $\lambda_r$ is selected via log-scale grid search on the validation split of the training dataset(s) within $[10^{-3}, 10^{-1}]$, and fix it for all experiments. Unless otherwise specified, we set $\lambda_{\mathrm{env}}=0.1$ and $\lambda_{\mathrm{cal}}=1.0$ to balance the structural prior and calibration distillation. We use the same small constant $\varepsilon$ for all numerical clamps throughout the paper. All experiments are conducted on a single NVIDIA RTX 3090 GPU (24GB).

\begin{table*}[t]
\centering
\caption{Quantitative results on NYUv2. We compare different scale-and-shift estimation strategies across models and training settings under a unified evaluation protocol. Global optimizes a single scale and shift for the entire dataset, while Image-based methods estimate them per image. Median and Linear Fit leverage ground-truth depth statistics. Best results in each column are shown in bold.}
\label{tab:nyu}
\setlength{\tabcolsep}{5pt}
\renewcommand{\arraystretch}{0.95}
\resizebox{\textwidth}{!}{%
\begin{tabular}{l|c|c|ccc|ccc}
\toprule
Model & Scaling & Dataset
& $\delta < 1.25 \uparrow$
& $\delta < 1.25^2 \uparrow$
& $\delta < 1.25^3 \uparrow$
& Abs Rel $\downarrow$
& $\log_{10} \downarrow$
& RMSE $\downarrow$ \\
\midrule
ZoeDepth & Image & NYUv2
& 0.951 & 0.994 & 0.999 & 0.077 & 0.033 & 0.282 \\
\midrule

\multirow{8}{*}{DPT-Hybrid}
& Global & NYUv2
& 0.904 & 0.988 & \textbf{0.998} & 0.109 & 0.045 & 0.357 \\
& Image & NYUv2
& 0.914 & 0.990 & \textbf{0.998} & 0.097 & 0.042 & 0.350 \\
& \cellcolor{fitblue}Linear Fit & \cellcolor{fitblue}NYUv2
& \cellcolor{fitblue}0.926 & \cellcolor{fitblue}0.991 & \cellcolor{fitblue}0.999
& \cellcolor{fitblue}0.094 & \cellcolor{fitblue}0.040 & \cellcolor{fitblue}0.332 \\
& \cellcolor{methodgray}RSA & \cellcolor{methodgray}NYUv2
& \cellcolor{methodgray}0.916 & \cellcolor{methodgray}0.990 & \cellcolor{methodgray}\textbf{0.998}
& \cellcolor{methodgray}0.097 & \cellcolor{methodgray}0.042 & \cellcolor{methodgray}0.347 \\
& \cellcolor{methodgray}Ours & \cellcolor{methodgray}NYUv2
& \cellcolor{methodgray}\textbf{0.919} & \cellcolor{methodgray}\textbf{0.991} & \cellcolor{methodgray}\textbf{0.998}
& \cellcolor{methodgray}\textbf{0.095} & \cellcolor{methodgray}\textbf{0.041} & \cellcolor{methodgray}\textbf{0.342} \\
& Image & NYUv2,KITTI
& 0.911 & \textbf{0.989} & \textbf{0.998} & 0.098 & 0.043 & 0.355 \\
& \cellcolor{methodgray}RSA & \cellcolor{methodgray}NYUv2,KITTI
& \cellcolor{methodgray}0.913 & \cellcolor{methodgray}0.988 & \cellcolor{methodgray}\textbf{0.998}
& \cellcolor{methodgray}0.099 & \cellcolor{methodgray}0.042 & \cellcolor{methodgray}0.352 \\
& \cellcolor{methodgray}Ours & \cellcolor{methodgray}NYUv2,KITTI
& \cellcolor{methodgray}\textbf{0.916} & \cellcolor{methodgray}\textbf{0.989} & \cellcolor{methodgray}\textbf{0.998}
& \cellcolor{methodgray}\textbf{0.097} & \cellcolor{methodgray}\textbf{0.041} & \cellcolor{methodgray}\textbf{0.349} \\
\midrule

\multirow{7}{*}{DepthAnything-S}
& Image & NYUv2
& 0.749 & 0.965 & \textbf{0.997} & 0.169 & 0.068 & 0.517 \\
& \cellcolor{fitblue}Linear Fit & \cellcolor{fitblue}NYUv2
& \cellcolor{fitblue}0.965 & \cellcolor{fitblue}0.993 & \cellcolor{fitblue}0.997
& \cellcolor{fitblue}0.058 & \cellcolor{fitblue}0.025 & \cellcolor{fitblue}0.232 \\
& \cellcolor{methodgray}RSA & \cellcolor{methodgray}NYUv2
& \cellcolor{methodgray}0.775 & \cellcolor{methodgray}0.975 & \cellcolor{methodgray}\textbf{0.997}
& \cellcolor{methodgray}0.147 & \cellcolor{methodgray}0.065 & \cellcolor{methodgray}0.484 \\
& \cellcolor{methodgray}Ours & \cellcolor{methodgray}NYUv2
& \cellcolor{methodgray}\textbf{0.781} & \cellcolor{methodgray}\textbf{0.977} & \cellcolor{methodgray}\textbf{0.997}
& \cellcolor{methodgray}\textbf{0.144} & \cellcolor{methodgray}\textbf{0.064} & \cellcolor{methodgray}\textbf{0.476} \\
& Image & NYUv2,KITTI
& 0.710 & 0.947 & 0.992 & 0.181 & 0.075 & 0.574 \\
& \cellcolor{methodgray}RSA & \cellcolor{methodgray}NYUv2,KITTI
& \cellcolor{methodgray}0.776 & \cellcolor{methodgray}0.974 & \cellcolor{methodgray}\textbf{0.996}
& \cellcolor{methodgray}0.148 & \cellcolor{methodgray}0.065 & \cellcolor{methodgray}0.498 \\
& \cellcolor{methodgray}Ours & \cellcolor{methodgray}NYUv2,KITTI
& \cellcolor{methodgray}\textbf{0.782} & \cellcolor{methodgray}\textbf{0.976} & \cellcolor{methodgray}\textbf{0.996}
& \cellcolor{methodgray}\textbf{0.145} & \cellcolor{methodgray}\textbf{0.064} & \cellcolor{methodgray}\textbf{0.489} \\
\bottomrule
\end{tabular}
}
\end{table*}
\begin{table*}[t]
\centering
\caption{Quantitative results on the KITTI Eigen Split. The evaluation follows the same estimation protocol as in Table~\ref{tab:nyu}, focusing on cross-dataset generalization performance. Best results in each column are shown in bold.}
\label{tab:kitti}
\setlength{\tabcolsep}{5pt}
\renewcommand{\arraystretch}{0.95}
\resizebox{\textwidth}{!}{%
\begin{tabular}{l|c|c|ccc|ccc}
\toprule
Model & Scaling & Dataset
& $\delta < 1.25 \uparrow$
& $\delta < 1.25^2 \uparrow$
& $\delta < 1.25^3 \uparrow$
& Abs Rel $\downarrow$
& $\mathrm{RMSE}_{\log} \downarrow$
& RMSE $\downarrow$ \\
\midrule
ZoeDepth & Image & KITTI
& 0.971 & 0.996 & 0.999 & 0.054 & 0.082 & 2.281 \\
\midrule

\multirow{7}{*}{DPT-Hybrid}
& Image & KITTI
& 0.961 & \textbf{0.995} & \textbf{0.999} & 0.064 & 0.092 & 2.379 \\
& \cellcolor{fitblue}Linear Fit & \cellcolor{fitblue}KITTI
& \cellcolor{fitblue}0.974 & \cellcolor{fitblue}0.997 & \cellcolor{fitblue}0.999
& \cellcolor{fitblue}0.052 & \cellcolor{fitblue}0.080 & \cellcolor{fitblue}2.198 \\
& \cellcolor{methodgray}RSA & \cellcolor{methodgray}KITTI
& \cellcolor{methodgray}0.963 & \cellcolor{methodgray}\textbf{0.995} & \cellcolor{methodgray}\textbf{0.999}
& \cellcolor{methodgray}0.061 & \cellcolor{methodgray}0.090 & \cellcolor{methodgray}2.354 \\
& \cellcolor{methodgray}Ours & \cellcolor{methodgray}KITTI
& \cellcolor{methodgray}\textbf{0.965} & \cellcolor{methodgray}\textbf{0.995} & \cellcolor{methodgray}\textbf{0.999}
& \cellcolor{methodgray}\textbf{0.060} & \cellcolor{methodgray}\textbf{0.089} & \cellcolor{methodgray}\textbf{2.330} \\
& Image & NYUv2,KITTI
& 0.956 & 0.989 & 0.993 & 0.066 & 0.098 & 2.477 \\
& \cellcolor{methodgray}RSA & \cellcolor{methodgray}NYUv2,KITTI
& \cellcolor{methodgray}0.962 & \cellcolor{methodgray}0.994 & \cellcolor{methodgray}0.998
& \cellcolor{methodgray}0.060 & \cellcolor{methodgray}0.089 & \cellcolor{methodgray}2.342 \\
& \cellcolor{methodgray}Ours & \cellcolor{methodgray}NYUv2,KITTI
& \cellcolor{methodgray}\textbf{0.964} & \cellcolor{methodgray}\textbf{0.995} & \cellcolor{methodgray}\textbf{0.999}
& \cellcolor{methodgray}\textbf{0.059} & \cellcolor{methodgray}\textbf{0.088} & \cellcolor{methodgray}\textbf{2.320} \\
\midrule

\multirow{7}{*}{DepthAnything-S}
& Image & KITTI
& 0.768 & 0.951 & 0.983 & 0.162 & 0.195 & 4.483 \\
& \cellcolor{fitblue}Linear Fit & \cellcolor{fitblue}KITTI
& \cellcolor{fitblue}0.824 & \cellcolor{fitblue}0.896 & \cellcolor{fitblue}0.922
& \cellcolor{fitblue}0.149 & \cellcolor{fitblue}0.224 & \cellcolor{fitblue}3.595 \\
& \cellcolor{methodgray}RSA & \cellcolor{methodgray}KITTI
& \cellcolor{methodgray}0.780 & \cellcolor{methodgray}0.958 & \cellcolor{methodgray}0.988
& \cellcolor{methodgray}0.160 & \cellcolor{methodgray}0.189 & \cellcolor{methodgray}4.437 \\
& \cellcolor{methodgray}Ours & \cellcolor{methodgray}KITTI
& \cellcolor{methodgray}\textbf{0.787} & \cellcolor{methodgray}\textbf{0.961} & \cellcolor{methodgray}\textbf{0.989}
& \cellcolor{methodgray}\textbf{0.156} & \cellcolor{methodgray}\textbf{0.186} & \cellcolor{methodgray}\textbf{4.360} \\
& Image & NYUv2,KITTI
& 0.718 & 0.943 & 0.979 & 0.171 & 0.211 & 4.456 \\
& \cellcolor{methodgray}RSA & \cellcolor{methodgray}NYUv2,KITTI
& \cellcolor{methodgray}0.756 & \cellcolor{methodgray}0.956 & \cellcolor{methodgray}0.987
& \cellcolor{methodgray}0.158 & \cellcolor{methodgray}0.191 & \cellcolor{methodgray}4.457 \\
& \cellcolor{methodgray}Ours & \cellcolor{methodgray}NYUv2,KITTI
& \cellcolor{methodgray}\textbf{0.763} & \cellcolor{methodgray}\textbf{0.959} & \cellcolor{methodgray}\textbf{0.988}
& \cellcolor{methodgray}\textbf{0.154} & \cellcolor{methodgray}\textbf{0.188} & \cellcolor{methodgray}\textbf{4.380} \\
\bottomrule
\end{tabular}
}
\end{table*}

\subsection{Quantitative comparison}
Table~\ref{tab:nyu} and Table~\ref{tab:kitti} report in-domain results on NYUv2 and KITTI. Under the same frozen-backbone protocol, our calibration improves metric accuracy over the strongest baselines across both backbones, confirming more reliable global scale alignment.

\subsection{Qualitative comparison}
Figure~\ref{fig:qual_nyu} and Figure~\ref{fig:qual_kitti} illustrate the effect of calibration on metric scale. Our method reduces global scale mismatch and lowers absolute errors, especially in distant regions and low-texture surfaces.

\subsection{Zero-shot Transfer to Unseen Datasets}
We evaluate zero-shot transfer on SUN-RGBD and DDAD without fine-tuning, directly applying calibration heads trained on NYUv2 and KITTI. As shown in Table~\ref{tab:sun_rgbd} and Table~\ref{tab:ddad}, our method consistently outperforms vision-only and language-only calibration under domain shift, suggesting that a caption-conditioned envelope provides transferable scale constraints while vision-conditioned selection adapts calibration per image.

\begin{table}[t]
\centering
\caption{Zero-shot generalization on SUN-RGBD (indoor).}
\label{tab:sun_rgbd}
\setlength{\tabcolsep}{4.5pt}
\renewcommand{\arraystretch}{1.10}
\resizebox{\columnwidth}{!}{%
\begin{tabular}{l|c|c|ccc}
\toprule
Backbone & Scaling & Train
& $\delta < 1.25^3 \uparrow$
& Abs Rel $\downarrow$
& RMSE $\downarrow$ \\
\midrule

ZoeDepth-X & Image & NYUv2
& -- & 0.124 & 0.363 \\
\midrule

\multirow{5}{*}{DPT-Hybrid}
& Global & NYUv2, KITTI
& 0.984 & 0.154 & 0.482 \\
& Image & NYUv2, KITTI
& 0.984 & 0.153 & 0.478 \\
& \cellcolor{fitblue}Linear Fit & \cellcolor{fitblue}SUN-RGBD (oracle)
& \cellcolor{fitblue}0.993
& \cellcolor{fitblue}0.139
& \cellcolor{fitblue}0.412 \\
& \cellcolor{methodgray}RSA & \cellcolor{methodgray}NYUv2, KITTI
& \cellcolor{methodgray}\textbf{0.986}
& \cellcolor{methodgray}0.152
& \cellcolor{methodgray}0.463 \\
& \cellcolor{methodgray}Ours & \cellcolor{methodgray}NYUv2, KITTI
& \cellcolor{methodgray}\textbf{0.986}
& \cellcolor{methodgray}\textbf{0.147}
& \cellcolor{methodgray}\textbf{0.437} \\
\midrule

\multirow{4}{*}{DepthAnything-S}
& Image & NYUv2, KITTI
& 0.963 & 0.279 & 1.392 \\
& \cellcolor{fitblue}Linear Fit & \cellcolor{fitblue}SUN-RGBD (oracle)
& \cellcolor{fitblue}0.995
& \cellcolor{fitblue}0.113
& \cellcolor{fitblue}0.332 \\
& \cellcolor{methodgray}RSA & \cellcolor{methodgray}NYUv2, KITTI
& \cellcolor{methodgray}0.970
& \cellcolor{methodgray}0.238
& \cellcolor{methodgray}1.024 \\
& \cellcolor{methodgray}Ours & \cellcolor{methodgray}NYUv2, KITTI
& \cellcolor{methodgray}\textbf{0.975}
& \cellcolor{methodgray}\textbf{0.213}
& \cellcolor{methodgray}\textbf{0.967} \\
\bottomrule
\end{tabular}
}
\end{table}

\begin{table}[t]
\centering
\caption{Zero-shot generalization to DDAD (outdoor).}
\label{tab:ddad}
\setlength{\tabcolsep}{4.5pt}
\renewcommand{\arraystretch}{1.0}
\resizebox{\columnwidth}{!}{%
\begin{tabular}{l|c|c|ccc}
\toprule
Backbone & Scaling & Train
& $\delta < 1.25^3 \uparrow$
& Abs Rel $\downarrow$
& RMSE $\downarrow$ \\
\midrule

\multirow{5}{*}{DPT-Hybrid}
& Global & NYUv2, KITTI
& 0.969 & 0.183 & 15.967 \\
& Image & NYUv2, KITTI
& 0.975 & 0.179 & 14.468 \\
& \cellcolor{fitblue}Linear Fit & \cellcolor{fitblue}DDAD (oracle)
& \cellcolor{fitblue}0.990 & \cellcolor{fitblue}0.163 & \cellcolor{fitblue}10.342 \\
& \cellcolor{methodgray}RSA & \cellcolor{methodgray}NYUv2, KITTI
& \cellcolor{methodgray}0.981 & \cellcolor{methodgray}0.171 & \cellcolor{methodgray}13.539 \\
& \cellcolor{methodgray}Ours & \cellcolor{methodgray}NYUv2, KITTI
& \cellcolor{methodgray}\textbf{0.984}
& \cellcolor{methodgray}\textbf{0.167}
& \cellcolor{methodgray}\textbf{12.847} \\
\midrule

\multirow{5}{*}{DepthAnything-S}
& Global & NYUv2, KITTI
& 0.963 & 0.221 & 21.345 \\
& Image & NYUv2, KITTI
& 0.968 & 0.217 & 20.834 \\
& \cellcolor{fitblue}Linear Fit & \cellcolor{fitblue}DDAD (oracle)
& \cellcolor{fitblue}0.983 & \cellcolor{fitblue}0.182 & \cellcolor{fitblue}18.423 \\
& \cellcolor{methodgray}RSA & \cellcolor{methodgray}NYUv2, KITTI
& \cellcolor{methodgray}0.976 & \cellcolor{methodgray}0.207 & \cellcolor{methodgray}19.715 \\
& \cellcolor{methodgray}Ours & \cellcolor{methodgray}NYUv2, KITTI
& \cellcolor{methodgray}\textbf{0.979}
& \cellcolor{methodgray}\textbf{0.195}
& \cellcolor{methodgray}\textbf{19.054} \\
\bottomrule
\end{tabular}
}
\end{table}
\begin{table}[t]
\centering
\caption{Component ablation analysis on NYUv2 using the DPT-Hybrid backbone.}
\label{tab:ablation_components_nyu}
\resizebox{\columnwidth}{!}{%
    \setlength{\tabcolsep}{4pt} 
    \renewcommand{\arraystretch}{0.95}
    \begin{tabular}{ccc|ccc}
    \toprule
    \multicolumn{3}{c|}{Components} & \multicolumn{3}{c}{Metric Depth Metrics} \\
    Prior ($\boldsymbol{\mu}$) & Vis. ($\boldsymbol{\delta}$) & Env. ($\mathbf{r}$) & $\delta < 1.25 \uparrow$ & Abs Rel $\downarrow$ & RMSE $\downarrow$ \\
    \midrule
    -- & \checkmark & -- & 0.914 & 0.097 & 0.350 \\
    \rowcolor{methodgray}
    \checkmark & -- & -- & 0.915 & 0.097 & 0.349 \\
    \checkmark & \checkmark & -- & 0.915 & 0.096 & 0.351 \\
    \rowcolor{methodgray}
    \checkmark & \checkmark & \checkmark & \textbf{0.919} & \textbf{0.095} & \textbf{0.344} \\
    \bottomrule
    \end{tabular}%
}
\end{table}
\begin{table}[t]
\centering
\caption{Statistical analysis of predicted parameters across diverse captions.}
\label{tab:text_sensitivity_params}
\setlength{\tabcolsep}{8pt}
\renewcommand{\arraystretch}{1.0}
\begin{tabular}{l|cc|cc}
\toprule
\multirow{2}{*}{Method} & \multicolumn{2}{c|}{Log-Scale $\alpha$} & \multicolumn{2}{c}{Log-Shift $\beta$} \\
 & Mean & Std $\downarrow$ & Mean & Std $\downarrow$ \\
\midrule
Vision-only & -3.098 & \textbf{0.000} & -1.585 & \textbf{0.000} \\ 
\midrule
\rowcolor{methodgray}
Language-only (RSA-style) & -3.125 & 0.084 & -1.642 & 0.112 \\
\rowcolor{methodgray}
\textbf{Ours (Full)} & -3.104 & \textbf{0.052} & -1.602 & \textbf{0.065} \\ 
\bottomrule
\end{tabular}
\end{table}
\begin{table}[t]
\centering
\caption{Supervision ablation on NYUv2 using the DPT-Hybrid backbone.}
\label{tab:ablation_supervision_nyu}
\resizebox{\columnwidth}{!}{%
    \setlength{\tabcolsep}{8pt}
    \renewcommand{\arraystretch}{1.1}
    \begin{tabular}{lccc|ccc}
    \toprule
    Variant & $\mathcal{L}_{depth}$ & $\mathcal{L}_{env}$ & $\mathcal{L}_{cal}$ & $\delta < 1.25 \uparrow$ & Abs Rel $\downarrow$ & RMSE $\downarrow$ \\
    \midrule
    w/o Oracle & \checkmark & -- & -- & 0.908 & 0.103 & 0.358 \\
    \rowcolor{methodgray}
    w/o $\mathcal{L}_{cal}$ & \checkmark & \checkmark & -- & 0.914 & 0.098 & 0.349 \\
    w/o $\mathcal{L}_{env}$ & \checkmark & -- & \checkmark & 0.916 & 0.097 & 0.346 \\
    \rowcolor{methodgray}
    \textbf{Full (Ours)} & \checkmark & \checkmark & \checkmark & \textbf{0.919} & \textbf{0.095} & \textbf{0.342} \\
    \bottomrule
    \end{tabular}%
}
\end{table}

\subsection{Ablation Study}
\textbf{Effectiveness of Proposed Methods.} 
Table~\ref{tab:ablation_components_nyu} validates our envelope-based factorization $\tilde{\theta}=\mu(T)+r(T)\odot\delta(I)$. By constraining vision-guided selection within language-informed bounds, our method suppresses semantic noise and yields more consistent metric alignment than generic multimodal fusion.

\textbf{Impact of Training Strategy.} 
Table~\ref{tab:ablation_supervision_nyu} evaluates our hierarchical supervision. While $\mathcal{L}_{depth}$ alone (w/o Oracle) lacks explicit scale guidance ($0.103$ Abs Rel), $\mathcal{L}_{cal}$ leverages the closed-form oracle $\theta^*$ to linearize the inverse-depth relationship and significantly reduce error. Adding $\mathcal{L}_{env}$ further refines accuracy to $0.095$, proving that supervising both semantic bounds and geometric distillation is superior to optimizing for pixel-wise depth alone.

\textbf{Robustness to Different Text Input.}
We evaluate the sensitivity of calibration parameters to diverse captions for a fixed image; Table~\ref{tab:text_sensitivity_params} reports the mean and standard deviation (Std) across 15 different descriptions. Compared to language-only baselines, our method yields consistently lower variance, confirming that vision-guided selection stabilizes calibration against caption noise and ambiguity.

\section{Limitations}
Our method assumes that a frozen relative-depth backbone can be aligned to metric scale by a global affine calibration in inverse depth. While this low-cost design is effective, it may be insufficient when the backbone exhibits depth-dependent bias or local outliers (e.g., ``bowing'' or far-range compression), where a single mapping cannot recover high-fidelity metric depth. Extending the calibration beyond a single global fit (e.g., piecewise or locally varying transforms) could raise the performance ceiling, but would add degrees of freedom and depart from the frozen-backbone protocol studied here. In addition, we rely on auto-captions as a coarse prior; despite bounded instantiation, missing or incorrect semantics can still steer calibration.

\section{Conclusion}
We present a sensor-free option for recovering metric depth from a frozen relative-depth backbone by using language. Our method maps an automatically generated caption to an uncertainty envelope in an unconstrained global affine calibration space, and uses pooled multi-scale visual features to select a bounded, instance-specific calibration within that envelope. Across NYUv2 and KITTI, as well as zero-shot transfer to SUN-RGBD and DDAD, the proposed calibration consistently improves metric depth accuracy over existing language-guided baselines.

\bibliographystyle{IEEEtran}
\bibliography{refs}

\end{document}